\documentclass[letterpaper,10pt,conference]{ieeeconf}
\IEEEoverridecommandlockouts % This command is only needed if 
\usepackage[caption=false,font=normalsize,labelfont=sf,textfont=sf]{subfig}
\usepackage{url}
\usepackage{graphicx}
\usepackage{cite}
\makeatletter
\let\NAT@parse\undefined
\makeatother
\usepackage{amsmath}
\usepackage{amssymb}  % assumes amsmath package installed
\usepackage{hyperref}
\usepackage[skip=0.8pt]{caption}
\usepackage{multirow}
\usepackage{multicol}
\usepackage{booktabs}
\usepackage{color}
\usepackage{xcolor}
\usepackage[ruled,vlined,noend]{algorithm2e}
\providecommand{\keywords}[1]{\textbf{\textit{Index terms---}} #1}
\begin{document}
%
% paper title
% Titles are generally capitalized except for words such as a, an, and, as,
% at, but, by, for, in, nor, of, on, or, the, to and up, which are usually
% not capitalized unless they are the first or last word of the title.
% Linebreaks \\ can be used within to get better formatting as desired.
% Do not put math or special symbols in the title.
% \title{Title of the Paper}
% \title{Mimicking the Sonographer's Vision: Ultrasound Image Quality Assessment using Local-to-Global Anatomical Feature Extraction and Fusion}
% \title{\bf{Mimicking Sonographer's Vision: Local-to-Global Image Quality Assessment for Robotic Ultrasound}}
\title{\bf{Robotic Sonographer: Autonomous Robotic Ultrasound using Domain Expertise in Bayesian Optimization}}
\author{Deepak Raina$^{12*}$, SH Chandrashekhara$^3$, Richard Voyles$^2$, Juan Wachs$^2$, Subir Kumar Saha$^1$\\\vspace{-0.6cm}% <-this % stops a space % stops a space
% \thanks{We acknowledge the support for this work from Science and Engineering Research Board (India) and Purdue University (USA) - Overseas Visiting Doctoral Fellowship (Award No. SB/S9/Z-03/2017-VIII); Prime Minster's Research Fellowship (PMRF), IIT Delhi; and Daniel C. Lewis Professorship.}
\thanks{This work was supported in part by SERB (India) - OVDF Award No. SB/S9/Z-03/2017-VIII; PMRF - IIT Delhi under Ref. F.No.35-5/2017-TS.I:PMRF; National Science Foundation (NSF) USA under Grant \#2140612; Daniel C. Lewis Professorship and PU-IUPUI Seed Grant.}
\thanks{$^{1}$Indian Institute of Technology (IIT), Delhi, India (\{deepak.raina, saha\}@mech.iitd.ac.in); $^{2}$Purdue University (PU), Indiana, USA (\{draina, rvoyles, jpwachs\}@purdue.edu); $^{3}$All India Institute of Medical Sciences (AIIMS), Delhi, India (drchandruradioaiims@gmail.com).}
\thanks{$^{*}$Corresponding author is Deepak Raina}
% \thanks{$^{*}$Corresponding author Deepak Raina is a doctoral candidate at IIT Delhi and SERB-OVDF visiting scholar at Purdue University, USA. }
}
% IEEE PINS:
% Voyles: 101960; Saha: 102675; Chandru: 306017; Deepak: 226618; Wachs: 146275
% make the title area
% (Ref: F.No.35-5/2017-TS.I: PMRF)
\maketitle
% ############################
% ##### 200 WORDS ############
% ############################
\begin{abstract}
Ultrasound is a vital imaging modality utilized for a variety of diagnostic and interventional procedures. However, an expert sonographer is required to make accurate maneuvers of the probe over the human body while making sense of the ultrasound images for diagnostic purposes. This procedure requires a substantial amount of training and up to a few years of experience. In this paper, we propose an autonomous robotic ultrasound system that uses Bayesian Optimization (BO) in combination with the domain expertise to predict and effectively scan the regions where diagnostic quality ultrasound images can be acquired. The quality map, which is a distribution of image quality in a scanning region, is estimated using Gaussian process in BO. This relies on a prior quality map modeled using expert's demonstration of the high-quality probing maneuvers. The ultrasound image quality feedback is provided to BO, which is estimated using a deep convolution neural network model. This model was previously trained on database of images labelled for diagnostic quality by expert radiologists. Experiments on three different urinary bladder phantoms validated that the proposed autonomous ultrasound system can acquire ultrasound images for diagnostic purposes with a probing position and force accuracy of $\mathbf{98.7\%}$ and $\mathbf{97.8\%}$, respectively.

% The real-time assessment of ultrasound image quality is essential in the Robotic Ultrasound System (RUS) to acquire a diagnosable image. Our study of existing RUS systems revealed that they evaluated the quality of ultrasound images using pixel-level features such as confidence maps. However, in clinical examination, sonographers interpret the image quality based on its diagnostic value, which depends not only on pixel-level features but also on the combination of global appearance and local textural features. Our work utilizes this holistic approach of global and local image feature extraction and fusion to increase the diagnostic accuracy of ultrasound images acquired by RUS. We propose a deep learning model, USQNet, utilizing a Local-to-Global, Bilinear Pooling (L2G-BP) classifier for ultrasound image quality assessment. This classifier first extracts local and global information from ultrasound images using feature maps at multiple scales, and then uses a second-order pooling mechanism in order to exploit the statistical dependency of multi-scale features for fine-grained classification of image quality. We experimentally validated the USQNet for ultrasound images of the pelvic region with the subjective assessment by experienced radiologists. The results demonstrate that USQNet achieves an accuracy of $\boldsymbol{93\%}$, outperforms the confidence map methods by $\boldsymbol{85\%}$ and ablated versions of USQNet by $\boldsymbol{2-10\%}$. The project page with source code and dataset is available at \href{https://sites.google.com/view/usqnet}{\color{blue}\textit{{https://sites.google.com/view/usqnet}}}.
\end{abstract}

% Note that keywords are not normally used for peer review papers.
% \begin{keywords}
% Autonomous robotic ultrasound, Bayesian Optimisation, Expert's knowledge 
% \end{keywords}
% For peer review papers, you can put extra information on the cover
% page as needed:
% \ifCLASSOPTIONpeerreview
% \begin{center} \bfseries EDICS Category: 3-BBND \end{center}
% \fi
%
% For peerreview papers, this IEEEtran command inserts a page break and
% creates the second title. It will be ignored for other modes.
\IEEEpeerreviewmaketitle

% \vspace{-1.5mm}
\section{Introduction}
Ultrasound is the most frequently used imaging modality for diagnostic and surgical interventions due to its low cost, non-ionizing nature, portability and real-time feedback. Ultrasound offers several advantages over other imaging modalities, like Magnetic Resonance Imaging (MRI) and Computed Tomography (CT), however, the diagnosis by ultrasound is a highly operator-dependent modality \cite{rykkje2019hand}. This is because of the skills required for manual control of the probe and quality assessment of acquired images. Sonographers employ both directed as well as random explorations strategies to search for diagnostic-quality images. The ultrasound probe is moved within the region of interest through hand maneuvers initially and fine adjustments to the probe's translational and rotational motion later. These maneuvers also include the safe and precise adjustment of the pressure through the probe while simultaneously analyzing the quality of acquired images. Such an intricate procedure requires a great deal of skill, focus, experience and manual effort from sonographers. In rural settings, skilled sonographers availability is limited \cite{adams2021telerobotic}, and alternative solutions are required. 

\begin{figure}[t]
	\centering
	%trim={L,B,R,T}
	\includegraphics[trim=0cm 5cm 8cm 0cm,clip,width=\linewidth]{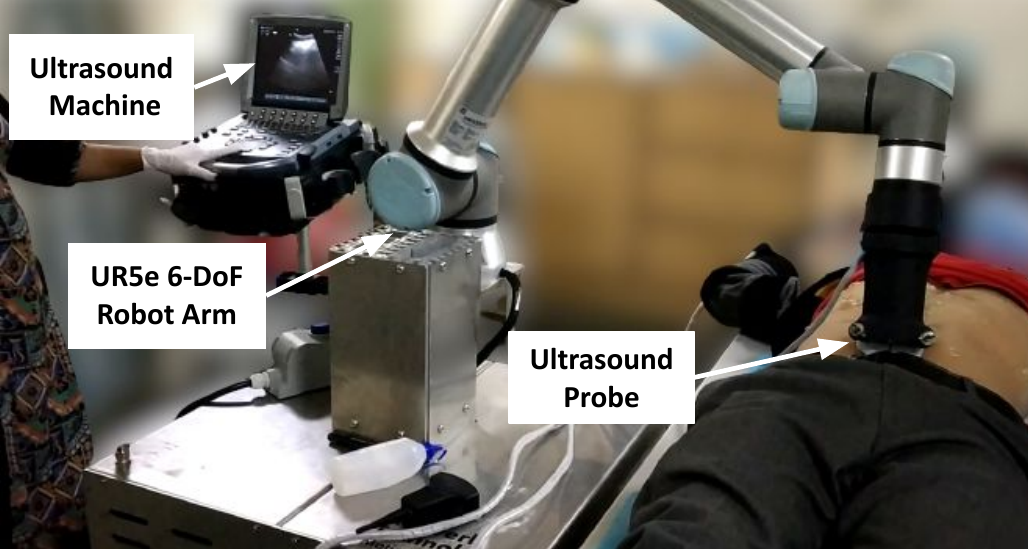}
	\caption{Robotic ultrasound system with probe attached to its end-effector \cite{raina2021comprehensive}, conducting a urinary bladder ultrasound.}
	\label{fig:real_exp_setup}
\end{figure}

In order to reduce the burden on experts, a Robotic Ultrasound System (RUS) is introduced. RUS consists of a dexterous robotic arm and an ultrasound machine with its probe attached to the end effector of the robot, as shown in Fig. \ref{fig:real_exp_setup}. RUS can help ensure the accuracy, safety and consistency of the ultrasound procedures. Recently, in order to address the aforementioned needs, several telerobotic or human-assisted ultrasound systems have been proposed \cite{raina2021comprehensive, wang2021application, duan20215g, adams2017initial, rojas2021assessing}.
Compared to these systems, a fully automated ultrasound system offers various potential benefits, including shorter procedure time, a shorter learning curve, minimal communication delays and a reduced cognitive load \cite{li2021overview}. However, there are key challenges for effective autonomous RUS. One
of the most important challenge has to do with the hand motions for ultrasound images acquisition. Such images exhibit considerable inter- and intra-subject variability and the image quality is highly dependent on the precise position, orientation and pressure of the ultrasound probe. With incorrect probe maneuvers, the resulting image presents noise, artifacts, blurred boundaries and poor visibility, thereby making it unacceptable for diagnosis. Sonographers rely on visual and haptic feedback, anatomical information, and diagnostic expertise from prior medical education to rapidly acquire the high-quality images.  Therefore, the RUS must locate the regions with acceptable diagnostic image quality for inter- and intra-patient procedures in the fewest exploration steps. 
% The idea of finding the region with high anatomical intensity is an optimisation problem to find the global maximum, where maxima is representing the high anatomical intensity region. 
% Sonographers often required 9 to 12 minutes to do an abdominal ultrasound which requires to scan multiple organs including liver, kidney, bladder and spleen \cite{raina2021comprehensive}.

% The objective of our system is to acquire high quality ultrasound images that a radiologist can analyze later for diagnostic assessment.
In this paper, we present an autonomous robotic ultrasound system that uses the domain-expertise in Bayesian Optimization (BO)-based search to scan the anatomical regions for acquiring diagnostic quality ultrasound images, thereby eliminating the need to thoroughly scan the entire region. The \textit{key contributions} of our work are as follows:
\begin{enumerate}
\item We proposed a prior in BO, gleaned from the expert's demonstration of high image quality probing poses, termed as \textit{expert's prior}. BO then estimates the region's unknown image quality as a semi-parametric Gaussian process model with expert's prior.
% We formulate a Bayesian optimization (BO) for autonomous robotic ultrasound in which the the unknown image quality map of the region is estimated as a semi-parametric residual Gaussian process model with a prior map obtained from expert's demonstration of high-quality probing locations.
\item A novel \textit{image quality metric} is proposed, trained using a dataset of ultrasound images labelled for diagnostic quality by expert radiologists, which provides image feedback of the region to the BO. 
% We integrated ultrasound image quality feedback of the region into the BO search, which is based on the image quality assessment model learned using a dataset of ultrasound images labelled for quality by an expert radiologist. 
\item We experimentally validated using three urinary bladder phantoms requiring different probing maneuvers for acquiring high image quality. The results show that our systems consistently and autonomously acquire high-quality ultrasound images in all phantoms.
\end{enumerate}
We believe that the use of BO combined with domain expertise to perform autonomous ultrasound scanning
will lead to less reliance on expert availability and a wider application in remote and underserved populations.
% To the best of our knowledge, this is the first attempt to use BO combined with domain-expertise to efficiently identify the region of interest for acquiring the ultrasound images with high diagnostic value.
\subsection{Related Work}
% Several attempts have been made by researchers worldwide to provide a framework for objective quality assessment of US images.
% \\
% \\
% \textbf{Quality assessment using Ultrasound image analysis:}
% \subsubsection{COn}
\noindent
\textbf{Autonomous Robotic Ultrasound Systems:}
In recent years, a range of autonomous robotic ultrasound systems has been proposed to minimize human intervention. Earlier works used image features for ultrasound image-based visual servoing \cite{mebarki20102, nakadate2011out, nadeau2016moments}. Later, various systems used pixel-based confidence map methods \cite{karamalis2012ultrasound} and segmentation of structures for optimizing the probe poses and forces \cite{chatelain2017confidence, jiang2020automatic, jiang2021autonomous, jiang2022towards}. However, these image feature- and pixel-based approaches are modality specific, computationally expensive and do not consider the significance of diagnostic aspects. Hennersperger \textit{et al.} \cite{hennersperger2016towards} developed the autonomous system using the pre-operative MRI scan, however, MRI is quite expensive to acquire. Ma \textit{et al.} \cite{ma2021autonomous} proposed autonomous lung scanning by localizing the target region using RGB-D sensor data. However, the system used only force feedback and did not rely on ultrasound image feedback, thereby limiting its diagnostic accuracy. 
% Hennersperger \textit{et al.} \cite{hennersperger2016towards} used the pre-operative MRI image to autonomously scan the abdominal region. Jiang \textit{et al.} \cite{jiang2022towards} proposed a vision-based approach using MRI data to autonomously scan the limb considering its articulated motion. Both of the

Recently, Li \textit{et al.} \cite{li2021autonomous, li2021image} proposed a deep Reinforcement Learning (RL) framework to control the probe for spinal ultrasound, incorporating image quality optimization into the reward formulation. However, the success of these systems is limited to phantoms and patients whose data was included during training. Moreover, deploying RL in medical systems is quite challenging, as it requires vast amount of physical interaction with the human body and poses safety and ethical concerns. In contrast to these systems, the proposed autonomous ultrasound system narrows down the area to be scanned using BO, eliminating the need to thoroughly scan the entire region. We further propose using domain expertise gleaned from the experts in the form of BO prior and image quality metrics, in order to acquire diagnostic-quality ultrasound images.
\\
\\
\textbf{Bayesian Optimization for Medical Robots:}
Due to the fast optimization capability, BO has been adopted for safety-critical robotic medical procedures, such as autonomous robotic palpation \cite{yan2021fast}, semi-autonomous surgical robot \cite{chen2020supervised}, controller tuning of hip exoskeletons \cite{ding2018human} and autonomous robotic ultrasound \cite{huang2021towards, goel2022autonomous}.
% Due to the fast optimization capability, BO has gained researchers' attention to use it for safety-critical robotic medical procedures, such as  Yan \textit{et al.} \cite{yan2021fast} used the BO for quick localization of tumor in the tissue via autonomous robotic palpation. Chen \textit{et al.} \cite{chen2020supervised} used the BO for semi-autonomous surgical robot in order to fine tune the user-specific role adaptation between human operator and robot. Ding \textit{et al.} \cite{ding2018human} used the BO to identify the control parameters of hip exoskeleton that minimizes the metabolic cost of walking. Goel \textit{et al.} \cite{goel2022autonomous} used BO for autonomous robotic ultrasound to find the region with high vessel density. 
Our work is a non-trivial extension to the work by Goel \textit{et al.} \cite{goel2022autonomous}. They proposed using BO for autonomous ultrasound utilizing segmentation of the vessel in the ultrasound image as feedback to the BO for scanning the region with high vessel density. They used hybrid position-force control to move the robot in $(x,y)$ plane while maintaining constant force along the $z-$direction to the point of contact. In contrast, our work suggests two technical improvements to enhance the practicality of this approach. First, we recommend using a deep learning model that generates quality scores for ultrasound images as feedback to the BO instead of relying on a segmented mask of the tissue or structure. The latter approach can be very time-consuming and labor-intensive for experts as they would need to annotate anatomical structures' boundaries, taking into account the ultrasound image noise and variability due to machine settings, probe pressure, and patient anatomy.
% First, we propose using the deep learning model providing ultrasound image quality scores as feedback to the BO. This is an alternative to using a segmented mask of the tissue or structure, which would be time-consuming and labor-intensive  for experts to annotate anatomical structures in the ultrasound image. This is  due to the image variability arising due to the differences in image quality, probe pressure, and anatomy of the patient.
Second, we expand the capabilities of the BO by enabling it to search for the optimal scanning region along the $(x,y,z)$-axis. Notably, the $z$-axis is under variable force control to account for varying physiological conditions \cite{akbari2021robotic}. 
% This adjustment in force is crucial for obtaining high-quality diagnostic images.
% Second, we extend the BO to search the optimal scanning region along the $(x,y,z)-$axis, where $z-$axis is under variable force control. The force adjustment is compulsory for acquiring diagnostic quality images from patients with different physiological conditions \cite{akbari2021robotic}.
% It often shows noise and is not always a good measure of diagnostic accuracy, which can cause BO to converge at local maxima (low diagnostic value) rather than global maxima.
%  The work in \cite{goel2022autonomous} used the segmentation of the ultrasound image to formulate the reward, however it is quite difficult to segment the regions of inter test in US images, which are quite noisy. However, the combined reward formulation will further enhance the performance of optimisation.
% \item In order to further reduce the number of steps to search the scanning region, we used the domain knowledge in the BO using the expert demonstrated trajectories of ultrasound probe. This knowledge has been manifested in Gaussian process priors that specify the initial beliefs on function to be approximated. The work in \cite{goel2022autonomous} used random prior for Gaussian process.
\\
\\
\textbf{Domain Expertise in BO:}
BO can utilize the expert's knowledge in the form of priors (beliefs) that the expert (practitioner) has on the potential location of the optimum. Such techniques have been mostly used for hyper-parameter tuning of image and text datasets \cite{wang2022pre}, open-source machine learning datasets \cite{hvarfner2022pi} and robot simulation experiments \cite{wang2018regret}. A few recent works have utilized expert's knowledge in the form of prior for medical robots \cite{ayvali2017utility,zhu2022automated}. Ayvali \textit{et al.} \cite{ayvali2017utility} propose robotic palpation to detect tissue abnormalities using BO. They modified the acquisition function of BO, whose value peaks at the user-provided locations. Zhu \textit{et al.} \cite{zhu2022automated} proposed an autonomous robotic auscultation system for locating the optimal sound quality location using BO. They used visual registration of the patient to locate the anatomical landmarks for obtaining a prior observation model. Inspired by these works, we propose BO for autonomous ultrasound leveraging a prior quality map gleaned from expert's demonstrations.
% To the best of our knowledge, the expert's knowledge in the form of BO prior has not yet been adopted for autonomous ultrasound scanning.
% Hvarfner et al. \cite{hvarfner2022pi} proposed a prior-weighted version of acquisition function for leveraging user-beliefs about the location of an optimum and demonstrated substantial performance gains on several vision-based deep-learning tasks. Wang et al. \cite{wang2022pre} manifests the prior knowledge by pre-training priors on different but related datsets. Wang et al. \cite{wang2018regret} obtain the estimate of prior from the dataset collected in offline phase. 
% \clearpage
\begin{figure*}[!ht]
	\centering
	%trim={L,B,R,T}
	\includegraphics[trim=0cm 2.5cm 0cm 0cm,clip,width=\linewidth]{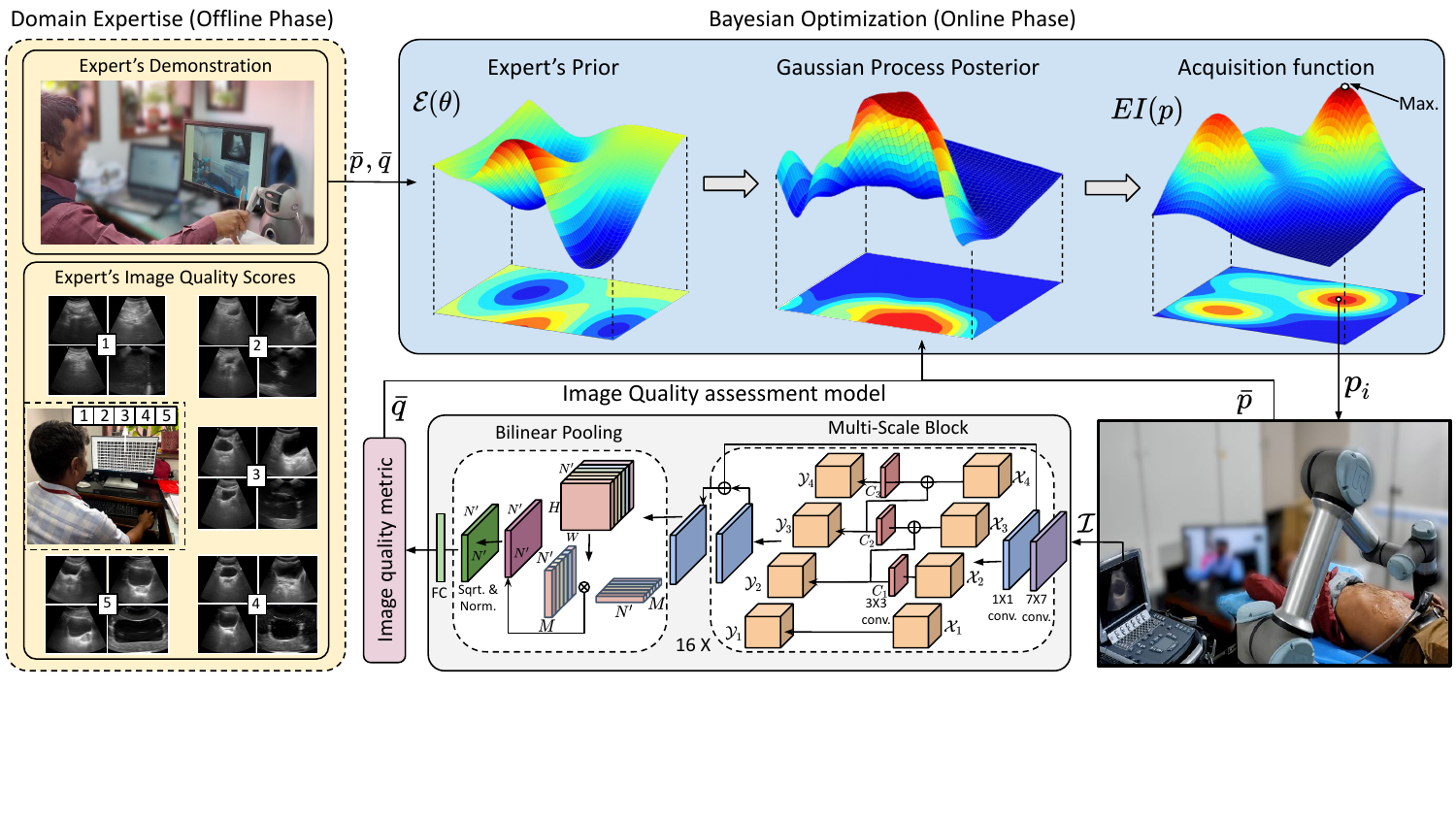}
	\caption{Overview of the pipeline for autonomous robotic ultrasound using online Bayesian optimization (BO), and offline domain expertise to obtain a prior quality map and to learn image quality assessment metric for providing feedback to BO.}
	\label{fig:overview}
\end{figure*}
% \vspace{-2mm}
\section{Methodology}
The pipeline of the autonomous robotic ultrasound system is shown in Fig. \ref{fig:overview}. In the \textit{offline phase}, the expert will demonstrate the potential probing poses to acquire the diagnostic quality images. This demonstrated data would be used to build a \textit{prior quality map}, which encodes prior anatomical approximation about expected image quality. We also built a dataset of urinary bladder ultrasound images of humans and phantoms with labelled image qualities and trained a deep learning model for image quality assessment metrics. In the \textit{online phase}, we used BO to select the probe poses to find the optimal ultrasound image quality utilizing both the prior map and quality metric gleaned from the domain expertise.
% which will explore the specified region of body to acquire the ultrasound image with high quality, representing the high diagnostic value of acquired image. The estimate of acquired image quality from the phantom has been termed as quality map. In order to quantify the quality of acquired image, we have learnt a model to mimic the expert radiologist's knowledge of ultrasound image quality analysis.

% \subsection{Model Architecture}
% The quantification of accurate quality of ultrasound image is challenging due to the presence of spurious image effects due to anatomical structure variations, blurred image boundaries, mirror artifacts, shadowing artifacts, and artifacts due to patient or sonographer's hand movement at the time of scanning. The attempts made in the literature for modeling the ultrasound image quality have lead to disappointing results when compared with radiologist's assessment. Therefore, we proposed a deep learning model to mimic the sonographer's assessment of ultrasound image quality via supervised learning.
\subsection{Bayesian optimization formulation}
We use BO to search adaptively for probing poses that yield a high-quality ultrasound image within a specified anatomical region. Let $A$ be the region of interest on the human body enclosing the anatomical structure, then the objective of BO is to solve:
% \vspace{-1em}
\begin{equation}
    \max_{\boldsymbol{p} \in A} q(\boldsymbol{\mathcal{I}}(\boldsymbol{p}))
\end{equation}
where $q(\boldsymbol{\mathcal{I}}(\boldsymbol{p}))$ denotes the quality score of ultrasound Image $\boldsymbol{\mathcal{I}}$ at probe pose $\boldsymbol{p}$. The BO will compute the probabilistic estimate of the unknown quality map $q(\boldsymbol{\mathcal{I}}(\boldsymbol{p}))$ across the human body using the domain expertise in the form of \textit{prior} and \textit{image quality metric}. An \textit{acquisition function} is optimized to yield the new probing pose. Once the new observation is found, the estimate is re-fitted to the data and the process is repeated till the termination criteria is reached, which is either the maximum reasonable iteration $N_{max}$ or the estimated quality score threshold required for adequate diagnosis. The overall algorithm is outlined in Algorithm 1.
\\
\vspace{-2mm}
\subsubsection{Expert's prior} \label{sec:pretrain}
A common estimator used in BO is Gaussian Process (GP) model, which defines an unknown function $f$ by assigning a probe pose $\boldsymbol{p}$ a random variable $f(\boldsymbol{p})$, which jointly represent a Gaussian. A GP for unknown function $f$ is defined by the mean function $\boldsymbol{\mu}(\cdot)$ and covariance or kernel function $\boldsymbol{\kappa}(\cdot,\cdot)$. Given the function value estimates $\bar{\boldsymbol{f}} = [{f}(\boldsymbol{p}_1), \cdots, {f}(\boldsymbol{p}_n)]$ at probe poses $\boldsymbol{\bar{p}} = [\boldsymbol{p}_1, \cdots, \boldsymbol{p}_n]$, GP regression can predict the function $f$ at new probe pose $\boldsymbol{p}^*$ as the Gaussian distribution and is given by:
\begin{equation}
    \mathcal{P}({f}(\boldsymbol{p}^*)|\boldsymbol{p}^*,\bar{\boldsymbol{p}}, \bar{\boldsymbol{f}}) = \mathcal{N}(\boldsymbol{k} \boldsymbol{K}^{-1} \bar{\boldsymbol{f}}, \boldsymbol{\kappa}(\boldsymbol{p}^*, \boldsymbol{p}^*) - \boldsymbol{k} \boldsymbol{K}^{-1} \boldsymbol{k}^T)
\end{equation}
where,
\begin{equation*}
\boldsymbol{k} = \begin{bmatrix}
\boldsymbol{\kappa}(\boldsymbol{p}_*, \boldsymbol{p}_1) & \cdots & \boldsymbol{\kappa}(\boldsymbol{p}_*, \boldsymbol{p}_n)
\end{bmatrix} 
\end{equation*}
\begin{equation*}
    \boldsymbol{K} = \begin{bmatrix}
\boldsymbol{\kappa}(\boldsymbol{p}_1, \boldsymbol{p}_1) & \cdots & \boldsymbol{\kappa}(\boldsymbol{p}_1, \boldsymbol{p}_n)\\
\vdots & \ddots & \vdots \\
\boldsymbol{\kappa}(\boldsymbol{p}_n, \boldsymbol{p}_1) & \cdots & \boldsymbol{\kappa}(\boldsymbol{p}_n, \boldsymbol{p}_n)
\end{bmatrix}
\end{equation*}
We opted to use a combination of two kernel functions, namely the radial basis function and white noise function, as their combination improved estimations for structures present in ultrasound images \cite{goel2022autonomous}. The formulation of the kernel is:
\begin{equation}
   {\kappa}(\boldsymbol{p}_i, \boldsymbol{p}_j) = \sigma_r \exp \bigg(\frac{-||\boldsymbol{p}_i - \boldsymbol{p}_j||^2}{2l^2}\bigg) + \sigma_w \boldsymbol{{I}}
\end{equation}
where $\sigma_r$ is the overall variance, $l$ is the length-scale, $\sigma_w$ is the variance of noise and $\boldsymbol{{I}}$ is the identity matrix. We further denote the set of image qualities as $\bar{\boldsymbol{q}} = [q_1, \cdots, {q}_n]$.
% Ref for GP as prior: https://deisenroth.cc/teaching/2019-20/linear-regression-aims/lecture_gaussian_processes.pdf

In GP, we propose using prior knowledge gleaned from expert's demonstrations to reduce the explorations and capture the variations of probe poses on the magnitude of ultrasound image quality corresponding to different human anatomy. Inspired from work in \cite{zhu2022automated}, we formulated the GP as a semi-parametric GP model, with its prior $\mathcal{E}({\boldsymbol{\theta}})$ modeled as a Gaussian process with latent parameters $\boldsymbol{\theta}$, representing the mean $\boldsymbol{\mu}_{\boldsymbol{\theta}}$ and covariance function $\boldsymbol{\kappa}$. The parameters $\boldsymbol{\theta}$ is initially inferred from observed probe poses and ultrasound image qualities, which the expert will provide by maneuvering the probe at the potential poses of the optimum image quality across different subjects. During online BO, $\boldsymbol{\theta}$ will be inferred using the history of points in $(\bar{\boldsymbol{p}}, \bar{\boldsymbol{q}})$ and prior $\mathcal{E}({\boldsymbol{\theta}})$ with Maximum A Posteriori (MAP) estimation, using an L-BFGS solver as:
% Limited memory Broyden–Fletcher–Goldfarb–Shanno (
% Although the large amount of exploration by BO will converge to the optimal in most cases, the expert's prior would capture the variations of probe poses corresponding to different human anatomy on the magnitude of ultrasound image quality. Therefore, we describe the expert's prior $E({\boldsymbol{\theta}})$ a distribution function of the latent variables $\boldsymbol{\theta} = [d_x, d_y, d_f, s_q]$, representing the position and foce offsets ($d_x, d_y, d_f$) and image quality scaling $s_q$ and infer the $\boldsymbol{\theta}$ from observed ultrasound image qualities. To obtain the prior of quality map, expert's will place the probe at the potential locations of the optimum image quality across different subjects. The prior will then be modeled using the multivariate Gaussian distribution $\mathcal{N}(\mu_{\boldsymbol{\theta}}, \Sigma)$, where $\boldsymbol{\theta}_0$ is the initial estimate of parameters. Then $\boldsymbol{\theta}$ will be inferred using the history of points in $\bar{\boldsymbol{x}}$ and $\bar{\boldsymbol{e}}$ with Maximum A Posteriori (MAP) estimation, using a Limited memory Broyden–Fletcher–Goldfarb–Shanno (L-BFGS) solver:
\vspace{-2mm}
\begin{equation}
    \boldsymbol{\theta}^* = \arg\max_{\boldsymbol{\theta}} \mathcal{L} (\boldsymbol{\theta}| \bar{\boldsymbol{p}}, \bar{\boldsymbol{q}}) \mathcal{E}(\boldsymbol{\theta})
\end{equation}
% \vspace{-2mm}
where $ \mathcal{L} (\boldsymbol{\theta}| \bar{\boldsymbol{p}}, \bar{\boldsymbol{q}}) = \prod \mathbb{P} (\boldsymbol{q}_i| \boldsymbol{\mu}_{\boldsymbol{\theta}}(\boldsymbol{p}_i), \boldsymbol{K})$ is the likelihood function and $\mathbb{P}(.)$ denotes the probability density function of Gaussian distribution $\mathcal{N}(\boldsymbol{q}_i| \boldsymbol{\mu}_{\boldsymbol{\theta}}(\boldsymbol{p}_i), \boldsymbol{K})$. Since GP models the residual function $f(\boldsymbol{p})$ with respect to the prior, we subtract the prior from image quality as $f(\boldsymbol{p}_i) =  q_i - {\mu}_{\boldsymbol{\theta}}(\boldsymbol{p})$, before re-estimating the GP.
\begin{algorithm}[h]
\SetAlgoLined
\textbf{Input:} Prior $\mathcal{E}{(\boldsymbol{\theta})}$, Region $A$, max. iterations $N_{max}$\;
Initialize $\Bar{\mathbf{p}} = \{\}$, $\Bar{\mathbf{f}} = \{\}$, $\Bar{\mathbf{q}} = \{\}$\;
  \For{$i = 1,...,N_{max}$ }{
    $\boldsymbol{p}_i \leftarrow \arg\max_{\boldsymbol{p} \in A} EI(\boldsymbol{p})$\;
    \eIf{termination criteria reached}{
        stop\;
    }{
        Probe at~$\boldsymbol{p}_i$, compute image quality $q_i$\;
        Set $\Bar{\mathbf{p}} \gets \Bar{\mathbf{p}} \cup \{p_i\}$, $\Bar{\mathbf{q}} \gets \Bar{\mathbf{q}} \cup \{q_i\}$\;
        $\boldsymbol{\theta} \gets \text{argmax} \, \mathcal{L}(\boldsymbol{\theta}|\Bar{\mathbf{p}},\Bar{\mathbf{q}}) \mathcal{E}(\boldsymbol{\theta})$\; 
        Set $\Bar{\mathbf{f}} \gets \Bar{\mathbf{f}} \cup \{ q_i - {\mu}_{\boldsymbol{\theta}}(\boldsymbol{p})\}$\;
        Re-estimate GP\;
    }
  }
 \Return Top probe poses with max. image quality\;
 \caption{Bayesian Optimization for Ultrasound}
 \label{alg:BO}
\end{algorithm}
\subsubsection{Acquisition Function} In each iteration of BO, the next probe pose to observe the image quality is determined using an acquisition function. We have used an Expected Improvement ($EI$), which is the most commonly used acquisition function. If the posterior mean and variance of GP is given by $\boldsymbol{\mu}_{\bar{\boldsymbol{f}}}(\boldsymbol{x}), \boldsymbol{\sigma}_{\bar{\boldsymbol{f}}}^2(x)$, then $EI$ can be formulated as:
\begin{equation} \label{eq:ei}
    \resizebox{0.9\hsize}{!}{$EI(\boldsymbol{p}) = 
    \begin{cases}
    (\boldsymbol{\mu}_{\bar{\boldsymbol{f}}}(\boldsymbol{p}) - f^+(\boldsymbol{p}) - \xi)\boldsymbol{\Phi}(\boldsymbol{Z}) +\boldsymbol{\sigma}_{\bar{\boldsymbol{f}}}^2(\boldsymbol{p})\boldsymbol{\phi}(\boldsymbol{Z}) & \text{if } \boldsymbol{\sigma}_{\bar{\boldsymbol{f}}}^2(\boldsymbol{p}) > 0\\
    0              & \text{if } \boldsymbol{\sigma}_{\bar{\boldsymbol{f}}}^2(\boldsymbol{p}) = 0
    \end{cases}$}
\end{equation}
% \begin{equation} \label{eq:ei}
%     EI(\boldsymbol{p}) = 
%     \begin{cases}
%     (\boldsymbol{\mu}_{\bar{\boldsymbol{f}}}(\boldsymbol{p}) - q^\dagger - \xi)\boldsymbol{\Phi}(\boldsymbol{Z}) +\boldsymbol{\sigma}_{\bar{\boldsymbol{f}}}^2(\boldsymbol{p})\boldsymbol{\phi}(\boldsymbol{Z}) & \text{if } \boldsymbol{\sigma}_{\bar{\boldsymbol{f}}}^2(\boldsymbol{p}) > 0\\
%     0              & \text{if } \boldsymbol{\sigma}_{\bar{\boldsymbol{f}}}^2(\boldsymbol{p}) = 0
%     \end{cases}
% \end{equation}
% \begin{equation}
%   \mathcal{L} =
%   \begin{cases}
%     1 & \text{if $i = j$ and $deg_j \neq 0 $} \\
%     -\frac{1}{\sqrt{deg_i deg_j}} & \text{if $(i, j) \in E$} \\
%     0 & \text{otherwise}
%   \end{cases}
% \end{equation}
% \begin{equation}
%     \resizebox{0.5\hsize}{!}{\boldsymbol{Z} = \begin{cases} \frac{\boldsymbol{\mu}_{\bar{\boldsymbol{f}}}(\boldsymbol{p}) - {f}(\boldsymbol{p}^+)}{\boldsymbol{\sigma}_{\bar{\boldsymbol{f}}}^2(p)} &  \text{if } \boldsymbol{\sigma}_{\bar{\boldsymbol{f}}}^2(\boldsymbol{p}) > 0\\
%     0 & \text{if } \boldsymbol{\sigma}_{\bar{\boldsymbol{f}}}^2(\boldsymbol{p}) = 0
%     \end{cases}}
% \end{equation}
where $\boldsymbol{Z} = \frac{\boldsymbol{\mu}_{\bar{\boldsymbol{f}}}(\boldsymbol{p}) - {f^+}(\boldsymbol{p})}{\boldsymbol{\sigma}_{\bar{\boldsymbol{f}}}^2(p)}$ if $\boldsymbol{\sigma}_{\bar{\boldsymbol{f}}}^2(\boldsymbol{p}) > 0$ else $0$; $\boldsymbol{\Phi}$ and $\boldsymbol{\phi}$ are the probability and cumulative density function of standard normal distribution, respectively and $f^+(\boldsymbol{p})$ is the best observed quality so far. The parameter $\xi$ in eq. \eqref{eq:ei} governs the amount of exploration during optimization and a high $\xi$ value leads to more exploration or less exploitation.
\subsection{Expert's ultrasound image quality metric} \label{sec:usiqa}
\subsubsection{Dataset} 
We used two datasets of Urinary Bladder (UB) ultrasound images. One of them is collected during the \textit{in-vivo} trials of our in-house developed Telerobotic Ultrasound System \cite{raina2021comprehensive, chandrashekhara2022robotic} at All India Institute of Medical Sciences (AIIMS), Delhi, India. The AIIMS ethics committee approved this study under IEC-855/04.09.2020,RP-16/2020. The other dataset is collected from the UB phantom. A total of $2016$ real and $2016$ phantom images were collected. The ground truth quality of the images is an average integer score of labels by three expert radiologists, each having $15$ years of experience in abdomen radiology. Each label is an integer score between $1-5$, based on an internationally prescribed generalized 5-level absolute assessment scale \cite{cantin2014diagnostic, duan20215g}. A score of $1$ means no appearance of the urinary bladder and $5$ means that the clear depiction of the urinary bladder with distinct boundaries and acceptable artifacts, depicting a high diagnostic accuracy. A subpar-quality image ($2$ to $4$) either contains noise or motion artifacts, blurred images, indistinct boundaries, obscuring the posterior or anterior sections of the urinary bladder. Later, we normalized the quality score in the range of $0-1$ for standard comparison with other quality estimation methods.
\subsubsection{Feature extraction} Ultrasound image quality assessment requires rich feature extraction for classifying the images that are highly variable in appearance but differ a lot in terms of image quality, as shown in Fig. \ref{fig:overview}. In recent work, Song \textit{et al.} \cite{song2022medical} proposed a bilinear Convolutional Neural Network (CNN) for fine-grained classification of breast ultrasound image quality. We propose a technical enhancement to this work in order to analyze the urinary bladder ultrasound images, in which the bladder appears at multiple scales/shapes (refer Fig. \ref{fig:overview} for sample images) due to the variability among inter- and intra-subject anatomy, probe poses and forces. Thus, it is also essential to analyze images at multiple scales. Recently, Basu \textit{et al.} \cite{basu2022surpassing} proposed combining multi-scale and second-order capability for detecting gall bladder cancer. Taking inspiration from these works, we proposed a deep CNN-based quality assessment model. The base network used is Residual Network (ResNet50) \cite{he2016deep} and combined with Multi-scale, Bilinear Pooling classifier, as shown in Fig. \ref{fig:overview}.

We used the group convolution kernels on equal-width feature volume splits in place of the $3 \times 3$ convolution kernel in the bottleneck layer of ResNet50.  If $\boldsymbol{\mathcal{X}} \in \mathbb{R}^{H \times W \times N}$ represents the feature volume with height $H$, width $W$ and number of channels $N$, then the operation of Multi-scale block can be represented by the following equations:
% \begin{equation} \label{eq1}
% \begin{split}
% \boldsymbol{\mathcal{Y}}_1 & = \boldsymbol{\mathcal{X}}_1,  \boldsymbol{\mathcal{Y}}_2 & = \boldsymbol{C}_1(\boldsymbol{\mathcal{X}}_2) \\
% \boldsymbol{\mathcal{Y}}_3 & = \boldsymbol{C}_2(\boldsymbol{\mathcal{Y}}_2 + \boldsymbol{\mathcal{X}}_3), 
% \boldsymbol{\mathcal{Y}}_4 & = \boldsymbol{C}_3(\boldsymbol{\mathcal{Y}}_3 + \boldsymbol{\mathcal{X}}_4)
% \end{split}
% \end{equation}
\begin{align*}
\boldsymbol{\mathcal{Y}}_1 &= \boldsymbol{\mathcal{X}}_1 &  \boldsymbol{\mathcal{Y}}_3 & = \boldsymbol{C}_2(\boldsymbol{\mathcal{Y}}_2 + \boldsymbol{\mathcal{X}}_3) \\
\boldsymbol{\mathcal{Y}}_2 & = \boldsymbol{C}_1(\boldsymbol{\mathcal{X}}_2) &  \boldsymbol{\mathcal{Y}}_4 & = \boldsymbol{C}_3(\boldsymbol{\mathcal{Y}}_3 + \boldsymbol{\mathcal{X}}_4)
\end{align*}
where $\boldsymbol{\mathcal{X}}_i \in \mathbb{R}^{H \times W \times N/4} $. Each split $\boldsymbol{\mathcal{X}}_i$ is first concatenated with output of previous split $\boldsymbol{\mathcal{Y}}_{i-1}$ and then fed to the $3 \times 3$ convolutional kernel $\boldsymbol{C}_i$ to produce an output $\boldsymbol{\mathcal{Y}}_i$. After passing the image through $16$ multi-scale blocks, the image feature volume  $\boldsymbol{\mathcal{X}} \in \mathbb{R}^{H \times W \times N}$ is passed through $1 \times 1$ convolution block to reduce the feature volume to $\boldsymbol{X} \in \mathbb{R}^{H \times W \times N^{'}}$. Then it is reshaped to a matrix $\boldsymbol{\mathcal{X}} \in \mathbb{R}^{M \times N^{'}}$ where $M = H \times W$. Later, a bilinear pooling is applied as:
% \begin{equation} \label{eq:bpop}
%     \boldsymbol{\mathcal{B}} = \frac{1}{N^{'}} (\boldsymbol{\mathcal{X}}\boldsymbol{\mathcal{X}^T}) + \epsilon \boldsymbol{I}
% \end{equation}
\begin{align}
    \boldsymbol{\mathcal{B}} = \frac{1}{N^{'}} (\boldsymbol{\mathcal{X}}\boldsymbol{\mathcal{X}^T}) + \epsilon \boldsymbol{I} \label{eq:bp_pooling}\\
    \boldsymbol{\mathcal{B}} \xleftarrow[]{}     \frac{\boldsymbol{\mathcal{B}}}{||\boldsymbol{\mathcal{B}}||_2}
    % \boldsymbol{\mathcal{B}} 
    \xleftarrow[]{} 
    sign(\boldsymbol{\mathcal{B}})\sqrt{|\boldsymbol{\mathcal{B}}|} 
 \label{eq:bp_sqrt}
\end{align}
where eq. \eqref{eq:bp_pooling} computes the outer product of feature volume, eq. \eqref{eq:bp_sqrt} will first perform the element-wise square-root and then the $l_2$ normalisation of the matrix $\boldsymbol{\mathcal{B}}$. Finally, the flattening of the feature map is done and then a fully connected layer is utilized to return the ultrasound image quality score.

% \subsubsection{Learning and testing}

\subsection{Robot control}
The robot controller will move the probe to the new pose $\boldsymbol{p} = [x, y, z]$ given by BO, where $(x, y)$ is under position control and $z$ is under force ($f_z$) control. For the safety of phantoms, the force limits have been set to $20 N$ \cite{raina2021comprehensive}. The orientation of the probe is kept normal to the point of contact. The hybrid position-force control is used for controlling the robot. After searching, the robot may execute the top probe poses with maximum image quality.
\section{Results and Discussions} \label{sec:results}
\subsection{Experimental setup} \label{sec:exp_setup}
\begin{figure}[!ht]
	\centering
	%trim={L,B,R,T}
	\includegraphics[trim=0cm 1cm 0cm 0cm,clip,width=0.9\linewidth]{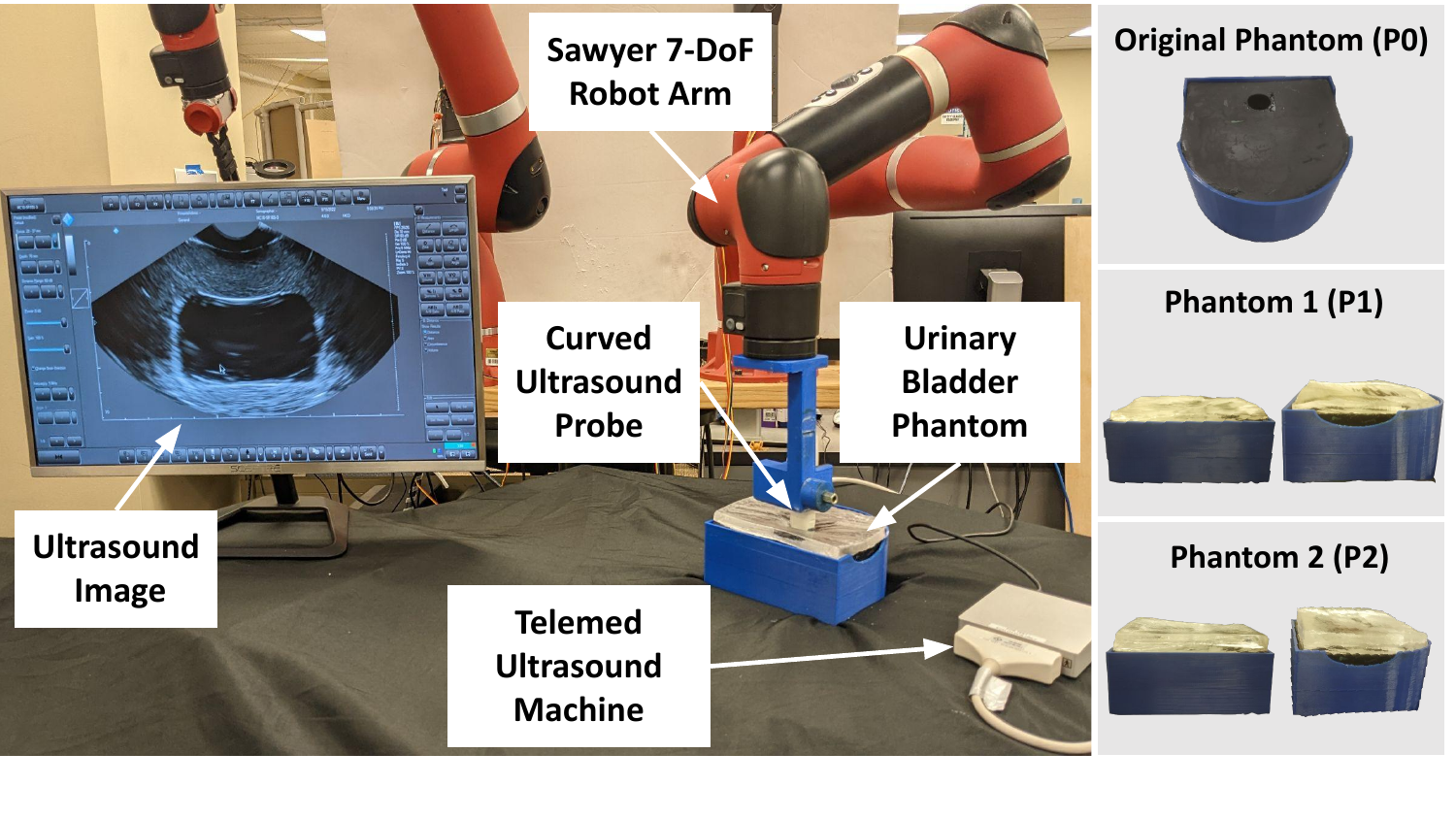}
	\caption{Experimental setup of robotic ultrasound system with three phantoms of the urinary bladder.}
	\label{fig:exp_setup}
\end{figure}

% \begin{figure}[!ht]
% 	\centering
% 	%trim={L,B,R,T}
% 	\includegraphics[trim=0cm 0cm 6cm 0cm,clip,width=0.8\linewidth]{figs/exp_setup2}
% 	\caption{Experimental setup of robotic ultrasound system.}
% 	\label{fig:exp_setup}
% \end{figure}
We conducted the experiments on the laboratory setup of the Robotic Ultrasound System at Purdue University, USA, consisting of a $7$-DoF Sawyer collaborative robotic arm (Rethink Robotics, Germany) with Micro Convex MC10-5R10S-3 transducer attached to its end-effector. The US image is captured by the Telemed Ultrasound machine (Telemed Medical Systems, Italy) and is transferred to the laptop.  The ultrasound was performed on a urinary bladder phantom (YourDesignMedical, USA). We customized this phantom with the $0.39$ inches thick (subjected to manual cutting error) rectangular layers of ballistic gel in order to approximately represent the patient's body with physiological differences. Thus, we present our results using three phantoms, termed as $P0$, $P1$ and $P2$, having $0$, $1$ and $2$ layers, as shown in Fig. \ref{fig:exp_setup}. 
The BO and image quality model have been implemented in Python $3.8$ and PyTorch $1.11$. ROS has been used to integrate and establish communication among all components of the setup. For BO Algorithm 1, we used $\xi=0.1$, $N_{max}=50$, $A \in ((0,0.15)m, (0,0.15)m, (8-20)N)$ for $(x,y,f_z)$. The prior $\mathcal{E}(\boldsymbol{\theta})$ has been modeled using GP by fitting it to $10$ potential probing poses and corresponding image qualities.
% The prior quality map described in Section \ref{sec:pretrain} has been build using demonstrations from P
% The Bayesian optimization and image quality assessment model has been implemented in Python $3.8$ and PyTorch $1.11$, respectively. Robot Operating System (ROS) has been used to integrate and establish communication among all components of the setup. For Bayesian optimisation, the value of parameter $\xi$ used in eq. \eqref{eq:ei} is $0.1$ to balance the exploration and exploitation. For GP, we used the observation noise $\alpha=0.004$. The maximum number of iterations $N_{max}$ in BO is used as $50$. BO in Algorithm 1 has been provided with the scanning surface limits and force limits $(8-20N)$.
% \subsection{Implementation Details}
% The quality assessment model has been implemented using PyTorch 1.11 in Python 3.8. The optimizer used for training is Stochastic Gradient Descent (SGD) with a learning rate of $0.001$, momentum of $0.9$ and weight decay of $0.0001$. The size of input US image is $224 \times 224$. The batch size used is $16$. The model was trained for $100$ epochs. The details of segmentation model is available in \cite{raina2022slimunet}. 
\subsection{Performance of quality assessment model}
We trained the ultrasound image quality assessment model, explained in Section \ref{sec:usiqa}, using the Categorical Cross Entropy (CCE) as a loss function. We split the dataset in a $90:10$ ratio as a training and testing dataset. We also used the transfer learning approach \cite{cheng2017transfer} and used the proposed model pre-trained on ImageNet. The stochastic gradient descent has been used as an optimizer with a learning rate of $0.005$, momentum of $0.9$ and weight decay of $0.0005$. The size of the input image to the model is $224 \times 224$, batch size is $16$ and the network is trained for $100$ epochs. The results in Table \ref{tab:usqnet_evaluate} shows that the proposed model (ResNet50+MS+BP) achieved an increase in accuracy by $3.01\%$ on a test set when compared to the ResNet50+BP model proposed in \cite{song2022medical}. 
% We also conducted cross-validation study to ensure the generalizability of the model. 
% The weighted sampling strategy has been used during training to handle the unequal distribution of images for each quality score.
% The proposed model performance is evaluated on a test set and 10-fold cross-validation for classifying the US image quality. The values of Precision, Recall, and Accuracy for the test set and 10-cross validation test are shown in Table \ref{tab:usqnet_evaluate}. The USQNet reported a maximum accuracy of $94.0\%$ and average accuracy of $92.6\%$.
% \begin{table}[ht]
% \begin{minipage}[b]{0.6\linewidth}
% \centering
% \includegraphics[width=0.9\textwidth]{figs/cfm_l2_a92.png}
% \captionof{figure}{Confusion matrix}
% \label{fig:image}
% \end{minipage}%
% \begin{minipage}[b]{0.4\linewidth}
% \centering
% \resizebox{\textwidth}{!}{\begin{tabular}{cccc}
%     \hline
%     Score & Precision & Recall & F1-score \\
%     \midrule
%     $1$ & $98$ & $100$ & $99$ \\
%     $2$ & $100$ & $97$ & $98$ \\
%     $3$ & $90$ & $90$ & $90$ \\
%     $4$ & $92$ & $92$ & $92$ \\
%     $5$ & $95$ & $95$ & $95$ \\
%     \midrule
%     \textbf{Avg.} & $95$ & $95$ & $95$ \\
%     \bottomrule
%     \end{tabular}}
%     \caption{Test Set evaluation metrics of USQNet}
%     \label{table:student}
% \end{minipage}
% \end{table}
\renewcommand{\arraystretch}{1.1}
\begin{table}[ht]
  \centering
    \caption{Comparison of ultrasound image quality assessment model predictions on testing set, where accuracy values close to $100$ indicate similarity to the expert's quality score.}
  \resizebox{\linewidth}{!}{\begin{tabular}{ccccccc}
    \toprule
    \multirow{2}{*}{{\parbox{1cm}{\centering\textbf{Image quality score}}}}  & \multicolumn{3}{c}{\textbf{ResNet50+BP \cite{song2022medical}}} &  \multicolumn{3}{c}{\textbf{ResNet50+MS+BP (Proposed)}} \\
    % \cmidrule{2-8}
    \cmidrule(lr){2-4}\cmidrule(lr){5-7}
     & \textbf{Precision} & \textbf{Recall} & \textbf{Accuracy} & \textbf{Precision} & \textbf{Recall} & \textbf{Accuracy} \\
    \midrule
    % $\boldsymbol{1}$ & $98.1$ & $100.0$ & $99.1$ &  $98.7 \pm 1.6$ &  $95.1 \pm 3.0$ & $96.8 \pm 1.3$ \\
    % $\boldsymbol{2}$ & $100.0$ & $97.6$ & $98.8$ &  $88.4 \pm 5.1$ &  $94.1 \pm 3.5$ & $91.0 \pm 3.0$ \\
    % $\boldsymbol{3}$ & $90.4$ & $90.4$ & $90.4$ &  $87.4 \pm 5.4$ &  $82.5 \pm 12.7$ & $84.2 \pm 7.0$ \\
    % $\boldsymbol{4}$ & $92.6$ & $92.1$ & $92.4$ &  $91.3 \pm 3.9$ &  $92.2 \pm 1.9$ & $91.7 \pm 1.9$ \\
    % $\boldsymbol{5}$ & $95.2$ & $95.0$ & $95.1$ &  $94.2 \pm 2.4$ &  $94.4 \pm 2.9$ & $94.2 \pm 2.0$ \\
    % $\boldsymbol{1}$ & $97.92$ & $97.92$ & $97.92$ &  $98.7 \pm 1.6$ &  $95.1 \pm 3.0$ & $96.8 \pm 1.3$ \\
    % $\boldsymbol{2}$ & $93.33$ & $96.55$ & $94.91$ &  $88.4 \pm 5.1$ &  $94.1 \pm 3.5$ & $92.0 \pm 3.0$ \\
    % $\boldsymbol{3}$ & $89.47$ & $85.00$ & $87.18$ &  $87.4 \pm 5.4$ &  $82.5 \pm 12.7$ & $84.2 \pm 7.0$ \\
    % $\boldsymbol{4}$ & $91.84$ & $91.84$ & $91.84$ &  $91.3 \pm 3.9$ &  $92.2 \pm 1.9$ & $91.7 \pm 1.9$ \\
    % $\boldsymbol{5}$ & $94.87$ & $94.87$ & $94.87$ &  $94.2 \pm 2.4$ &  $94.4 \pm 2.9$ & $94.2 \pm 2.0$ \\
    % $\boldsymbol{1}$ & $98.56$ & $98.56$ & $98.56$ &  $99.1 \pm 1.2$ &  $95.9 \pm 2.7$ & $97.1 \pm 1.1$ \\
    % $\boldsymbol{2}$ & $94.16$ & $96.89$ & $93.91$ &  $90.4 \pm 4.6$ &  $93.9 \pm 3.3$ & $93.6 \pm 2.0$ \\
    % $\boldsymbol{3}$ & $88.77$ & $85.49$ & $87.01$ &  $86.2 \pm 5.9$ &  $81.1 \pm 13.3$ & $83.5 \pm 8.5$ \\
    % $\boldsymbol{4}$ & $92.05$ & $90.45$ & $92.33$ &  $92.3 \pm 1.3$ &  $91.1 \pm 1.5$ & $92.2 \pm 1.1$ \\
    % $\boldsymbol{5}$ & $95.33$ & $93.25$ & $92.16$ &  $93.6 \pm 1.6$ &  $94.5 \pm 2.6$ & $94.1 \pm 2.6$ \\
    $\boldsymbol{1}$ & $92.00$ & $97.87$ & $94.84$ &  $93.88$ &  $97.87$ & $95.88$ \\
    $\boldsymbol{2}$ & $83.33$ & $68.96$ & $75.47$ &  $83.87$ &  $89.65$ & $86.67$ \\
    $\boldsymbol{3}$ & $65.22$ & $75.00$ & $69.77$ &  $88.24$ &  $75.00$ & $81.08$ \\
    $\boldsymbol{4}$ & $93.33$ & $85.71$ & $89.36$ &  $91.11$ &  $83.67$ & $87.23$ \\
    $\boldsymbol{5}$ & $90.48$ & $97.44$ & $93.83$ &  $90.48$ &  $97.44$ & $93.83$ \\
    \midrule
    \textbf{Average} & $\boldsymbol{87.67}$ & $\boldsymbol{87.48}$ & $\boldsymbol{87.34}$ &  $\boldsymbol{90.23}$ &  $\boldsymbol{90.17}$ & $\boldsymbol{90.05}$ \\
    \bottomrule
  \end{tabular}}
    % \caption{The performance of USQNet on the test set and 10-fold cross-validation (Mean$\pm$SD).}
  \label{tab:usqnet_evaluate}
\end{table}
% \subsection{Validation of image segmentation network}
\subsection{Comparing different BO strategies}
\begin{figure*}[ht]
	\centering
	%trim={L,B,R,T}
	\includegraphics[trim=0cm 7cm 0cm 0cm,clip,width=0.9\linewidth]{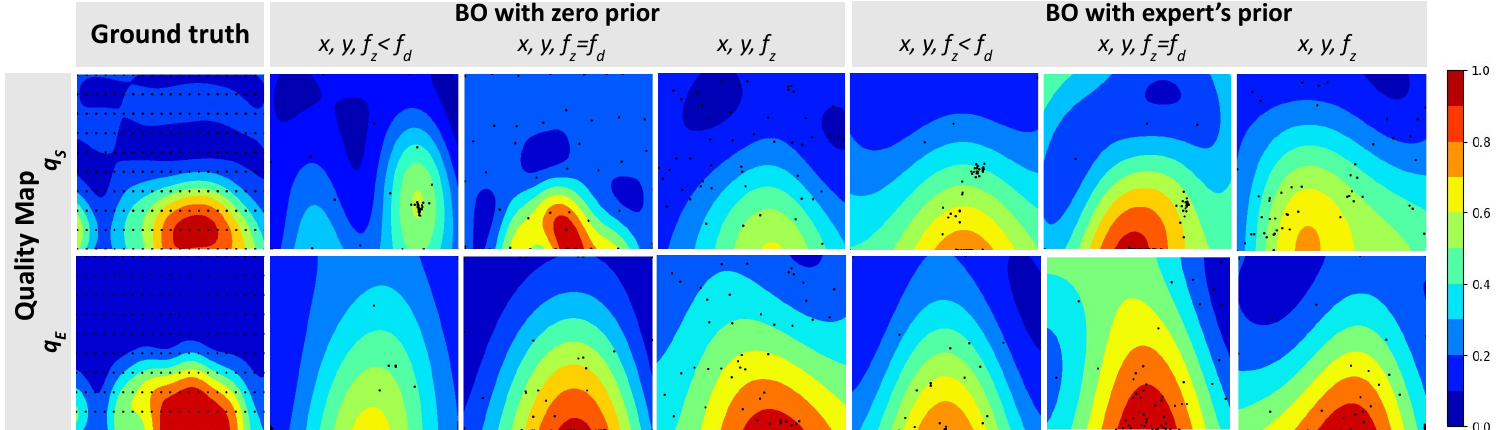}
	\caption{The estimated ultrasound image quality map of urinary bladder phantom $P0$ using different BO strategies. Black dots are positions where probe evaluated the quality. The corresponding ultrasound images are available in the attached media.}
	\label{fig:bo_cases}
\end{figure*}

% During the ultrasound procedure, the objective of sonographer is to scan the region with high anatomical intensity. The sonographers can search this region quickly with so much amount of experience doing the ultrasound over the years. 
% Bayesian optimization is an optimization approach that balances this exploration and exploitation for quickly searching the optimal vale of black-box function. 
In order to analyze the effectiveness of the proposed methodology, we have compared the BO with zero prior to the BO with the proposed expert's prior. We illustrated these search strategies using the image feedback having a mean of the segmented mask of the bladder in the ultrasound image ($q_S$) as used in \cite{goel2022autonomous} and having proposed ultrasound image quality metric learned from expert's rating ($q_E$). For segmentation, we used a U-net-based segmentation model proposed in \cite{raina2022slimunet}. Further, each of the feedback strategies has been compared with different search spaces, first considering the probe motion along $x$ and $y$-axis of the phantom, second along $x$, $y$ and $z-$axis of the phantom, where $z-$axis is under the force control $(f_z)$. The estimated quality maps obtained using these strategies for $P0$ are shown in Fig. \ref{fig:bo_cases}, where red region shows the high-quality region and blue region shows the low-quality region. The black dots over the map represents the queried probe positions over the phantom during the optimization. The first column in Fig. \ref{fig:bo_cases} shows the quality map obtained using the uniform movement of the probe over the phantom, which has been considered as the approximate ground truth quality map. For both the quality types, the ground truth has been obtained using the approximate desired force ($f_d$) of $14N$, $16N$ and $18N$ for $P0$, $P1$ and $P2$, respectively, which gives the best image quality in these phantoms. We present results for $3$ cases to illustrate the effect of searching with appropriate force in these phantoms: (i) $f_z < f_d$: $fz$ is constant but equal to $f_d-4$, (ii) $f_z = f_d$ and (iii) when $f_z$ is variable.
% The quality maps obtained for the above search strategies is shown in Fig. \ref{fig:bo_cases}, where red region shows the high quality region and blue region shows the low  quality region. The black dots over the map represents the queried probe positions over the phantom during the optimization. 

We compared the quality maps of these strategies by doing quantitative analysis using three metrics: (i) Sum of quality difference of top $n$ points, (ii) Top quality, and (iii) Zero Normalized Cross Correlation (ZNCC), as shown in Table \ref{tab:bo_quant}. The numbers in the table represent the average value of the matrices for the $3$ tests on each phantom. These metrics have been computed with respect to the approximate ground truth for the phantom. The sum of the difference between the top $n-$points compares the quality of images acquired from the top-n highest quality values, top quality compares the highest value of image quality score and ZNCC evaluates the overall similarity of the acquired quality map during the search. The value of quality differences close to $0$, and top quality and ZNCC value close to $1$ indicates a better estimation of the quality map. The quality maps in Fig. \ref{fig:bo_cases} with less scattered probe points (less exploration) and more points in the high-quality region (red) represent a better search strategy. 
% The value of  The ground truth has been computed for the phantom by uniform motion of the probe, where red color corresponds to high anatomical density region and blue color corresponds to low anatomical density regions. TEDE compares how well regions with high anatomical density are detected in different search strategies with respect to approximate ground truth. 
\begin{table*}[ht]
  \centering
  \vspace{-2mm}
    \caption{Quantitative comparison of different BO strategies for three different urinary bladder phantoms $P0, P1$ and $P2$}
    \resizebox{\linewidth}{!}{\begin{tabular}{cccccccccccccc}
    \toprule
    \multirow{3}{*}{\parbox{2cm}{\centering\textbf{Image quality estimation method}}} & \multirow{3}{*}{\textbf{Variables}} & \multicolumn{6}{c}{\textbf{BO with zero prior}} & \multicolumn{6}{c}{\textbf{BO with expert's prior}} \\
    \cmidrule(lr){3-8}\cmidrule(lr){9-14}
    ~ & ~ & \multicolumn{4}{c}{\textbf{Sum of quality difference of $n$ points}} & {\textbf{Top}} & \textbf{ZNCC} & \multicolumn{4}{c}{\textbf{Sum of quality difference of $n$ points}} & {\textbf{Top}} & \textbf{ZNCC} \\
    ~ & ~ & $n=1$ & $n=5$ & $n=10$ & $n=20$ & \textbf{Quality} & $n=50$ & $n=1$ & $n=5$ & $n=10$ & $n=20$ & \textbf{Quality} & $n=50$ \\
    \midrule
    {Segmentation} & $x,y,f_z<f_d$ & $0.382$ & $0.944$ & $1.430$ & $1.931$ & $0.684$ & $0.689$ & $0.291$ & $0.692$ & $0.941$ & $1.205$ & $0.782$ & $0.782$ \\
     ($q_{S}$) & $x,y,f_z=f_d$ & $0.132$& $0.609$ & $0.963$ & $1.651$ & $0.911$ & $0.811$ & $0.103$ & $0.531$ & $0.785$ & $1.308$ & $0.911$ & $0.920$ \\
    \cite{raina2022slimunet} & $x,y,f_z$ & $0.396$ & $1.016$ & $1.919$ & $3.336$ & $\mathbf{0.711}$ & $\mathbf{0.733}$ & $0.404$ & $0.991$ & $1.379$ & $2.158$ & $\mathbf{0.799}$ & $\mathbf{0.801}$\\
    \hline
     Expert's image& $x,y,f_z<f_d$ & $0.280$& $0.370$ & $0.600$ & $1.030$ & $0.690$ & $0.717$ & $0.130$ & $0.290$ & $0.570$ & $1.010$ & $0.750$ & $0.817$ \\
    quality metric & $x,y,f_z=f_d$ & $0.120$& $0.240$ & $0.390$ & $0.710$ & $0.950$ & $0.876$ & $0.050$ & $0.090$ & $0.180$ & $0.970$ & $0.980$ & $0.959$ \\
    ({$q_{E}$}) & $x,y,f_z$ & $0.130$ & $0.270$ & $0.820$ & $1.320$ & $\mathbf{0.823}$ & $\mathbf{0.821}$ & $0.040$ & $0.280$ & $0.760$ & $1.600$ & $\mathbf{0.910}$ & $\mathbf{0.889}$\\
    \bottomrule
  \end{tabular}}
  \label{tab:bo_quant}
\end{table*}

From the result in Fig. \ref{fig:bo_cases} and Table \ref{tab:bo_quant}, it has been found that the BO using the segmented image as quality score in $(x,y)$ space with $f_z \leq f_d$ have resulted in being too exploratory (low ZNCC) with a lot of points spread over the low-quality region of the phantom. However, the quality maps obtained using the expert's quality metric of the image have fewer explorations, with most of the probe positions in the high-quality region of the phantom. Due to noise and shadows in the ultrasound image, the segmentation results are prone to errors, resulting in a large number of probe evaluations in low-quality regions, whereas expert's image quality score, which is based on the holistic assessment of the image, pinpoints the focus on anatomical structures rather than getting distracted by noise. The search strategies using $fz<f_d$ could not find the high-quality region and instead converged to the local maxima rather than the global maxima. However, with $f_z=f_d$, the high-quality regions have been acquired. When the quality region is searched using $f_z$ as a variable in BO with zero prior, the quality maps and top quality score show that the high-quality regions can be located with a varying force too, which is essential for in-human ultrasound procedures. However, the search is quite exploratory, reporting low ZNCC values of $0.733$ and $0.821$ for quality $q_S$ and $q_E$, respectively.
% The search using different search spaces further demonstrates that BO is capable to search the high quality regions with varying forces.
When the expert's prior is used, all BO strategies have significantly improved, including the search space with three variables ($x,y,f_z$). The exploration steps of BO usually increase as the search space dimension increases. However, BO with expert's prior reported a top quality of $0.910$ with a ZNCC score of $0.889$, which is $9.6\%$ and $7.6\%$ more than the BO with zero prior.

\subsection{Validating the convergence of probe positions and forces}
\begin{figure}[!ht]
	\centering
	%trim={L,B,R,T}
	\includegraphics[trim=0cm 1cm 0cm 0.5cm,clip,width=\linewidth]{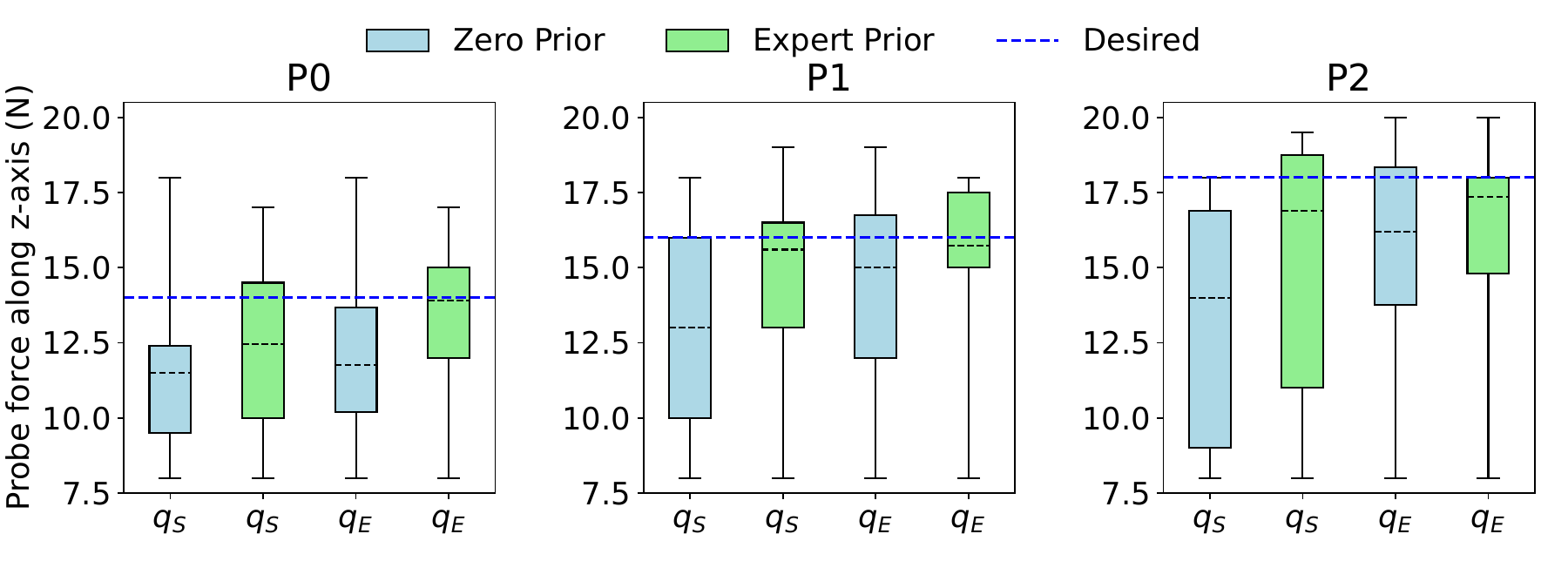}
	\caption{Force $f_z$ profiles with different BO search strategies}
	\label{fig:box_plots}
\end{figure}

Since our study involves phantom experiments, the approximate probe positions and forces that yield the best-quality images are known. The search strategy should converge to these approximate probe poses and forces to acquire high-quality images. The proposed strategy has reached the desired probe position with an average mean value accuracy of $98.73\%$ across all phantoms. To emphasize the convergence of force, we compared the probe forces explored by different BO search strategies, as shown in Fig. \ref{fig:box_plots}.
The proposed formulation of BO using the expert's prior and image quality metric has resulted in the mean value accuracy of $99.28\%$, $98.25\%$, and $96.11\%$ for $P0$, $P1$, and $P2$, respectively. Comparatively, the other BO search strategies using zero-prior and segmentation-based quality maps ($q_S$) have shown significant errors in mean values and greater standard deviation due to the noise in image feedback and the inability to adapt to the profile of the scanning region.
% In order to validate the convergence of algorithm in different cases, we have compared the Structural Similarity Index values and Peak signal noise ratio for image obtained at highest reward point with the the ground truth image of the phantom bladder in 10 cases. Also we compared the force and position convergence criteria for these cases. 
% KL Divergence - https://stats.stackexchange.com/questions/7440/kl-divergence-between-two-univariate-gaussians
\vspace{-2mm}
\section{Conclusion}
We proposed an autonomous Robotic Ultrasound System (RUS) to perform the ultrasound as per clinical protocols. We used Bayesian Optimization (BO) to search for high-quality regions leveraging the domain expertise in the form of a prior quality map and ultrasound image quality. The prior map has been gleaned using expert's demonstration of the potential high-quality probing maneuvers. A novel image quality metric has been learned from the expert-labelled dataset of ultrasound images. Three phantom experiments validated that incorporating domain expertise into BO effectively improves the system performance, resulting in acquiring diagnostic quality ultrasound images while adapting to desired probing maneuvers. Since phantom results are promising, we would like to validate its capability for \textit{in-vivo} study using our RUS in India \cite{raina2021comprehensive}, which is our future work. We would also expand the search space in BO from $[x,y,f_z]$ to include $[roll,pitch,yaw]$ in order to orient the probe for scanning patients with complex physiological conditions. 
\bibliography{references} 
\bibliographystyle{ieeetr}
% biography section
% If you have an EPS/PDF photo (graphicx package needed) extra braces are
% needed around the contents of the optional argument to biography to prevent
% the LaTeX parser from getting confused when it sees the complicated
% \includegraphics command within an optional argument. (You could create
% your own custom macro containing the \includegraphics command to make things
% simpler here.)
%\begin{IEEEbiography}[{\includegraphics[width=1in,height=1.25in,clip,keepaspectratio]{mshell}}]{Michael Shell}
% or if you just want to reserve a space for a photo:
% \begin{IEEEbiography}{Michael Shell}
% Biography text here.
% \end{IEEEbiography}
% % if you will not have a photo at all:
% \begin{IEEEbiographynophoto}{John Doe}
% Biography text here.
% \end{IEEEbiographynophoto}
% % insert where needed to balance the two columns on the last page with
% % biographies
% %\newpage
% \begin{IEEEbiographynophoto}{Jane Doe}
% Biography text here.
% \end{IEEEbiographynophoto}
% You can push biographies down or up by placing
% a \vfill before or after them. The appropriate
% use of \vfill depends on what kind of text is
% on the last page and whether or not the columns
% are being equalized.
%\vfill
% Can be used to pull up biographies so that the bottom of the last one
% is flush with the other column.
%\enlargethispage{-5in}
% that's all folks
\end{document}